\documentclass[sigconf]{acmart}
\usepackage{tabularx}   
\usepackage{array}      
\usepackage{booktabs}   
\usepackage{multirow}
\AtBeginDocument{%
  }

\setcopyright{acmlicensed}
\copyrightyear{2026}
\acmYear{2026}
\setcopyright{cc}
\setcctype{by}
\acmConference[WWW '26]{Proceedings of the ACM Web Conference 2026}{April 13--17, 2026}{Dubai, United Arab Emirates}
\acmBooktitle{Proceedings of the ACM Web Conference 2026 (WWW '26), April 13--17, 2026, Dubai, United Arab Emirates}
\acmPrice{}
\acmDOI{10.1145/3774904.3792434}
\acmISBN{979-8-4007-2307-0/2026/04}

\begin{document}

\title{EEO-TFV: Escape-Explore Optimizer for Web-Scale Time-Series Forecasting and Vision Analysis}

\author{Hua Wang}
\orcid{0000-0002-8844-9667}
\affiliation{%
	\institution{School of Computer and Artificial Intelligence, Ludong University}
	\city{Yantai}
	\state{Shandong}
	\country{China}
	\postcode{264025}
}
\email{hwa229@163.com}

\author{Jinghao Lu}
\orcid{0009-0003-9608-3380}
\affiliation{%
	\institution{School of Computer and Artificial Intelligence, Ludong University}
	\city{Yantai}
	\state{Shandong}
	\country{China}
	\postcode{264025}
}
\email{ljh@m.ldu.edu.cn}

\author{Fan Zhang}
\authornote{Corresponding author.}
\orcid{0000-0002-0343-3499}
\affiliation{%
	\institution{School of Computer Science and Technology, Shandong Technology and Business University}
	\city{Yantai}
	\state{Shandong}
	\country{China}
	\postcode{264005}
}
\email{zhangfan@sdtbu.edu.cn}

\begin{abstract}
	Transformer-based foundation models have achieved remarkable progress in tasks such as time-series forecasting and image segmentation. However, they frequently suffer from error accumulation in multivariate long-sequence prediction and exhibit vulnerability to out-of-distribution samples in image-related tasks. Furthermore, these challenges become particularly pronounced in large-scale Web data analysis tasks, which typically involve complex temporal patterns and multimodal features. This complexity substantially increases optimization difficulty, rendering models prone to stagnation at saddle points within high-dimensional parameter spaces. To address these issues, we propose a lightweight Transformer architecture in conjunction with a novel Escape-Explore Optimizer (EEO). The optimizer enhances both exploration and generalization while effectively avoiding sharp minima and saddle-point traps. Experimental results show that, in representative Web data scenarios, our method achieves performance on par with state-of-the-art models across 11 time-series benchmark datasets and the Synapse medical image segmentation task. Moreover, it demonstrates superior generalization and stability, thereby validating its potential as a versatile cross-task foundation model for Web-scale data mining and analysis.
\end{abstract}


\maketitle

\section{Introduction}

In recent years, Transformer-based foundation models have achieved remarkable breakthroughs across diverse domains, including image segmentation and time-series forecasting \cite{li2023time, lu2025mcnr}. Owing to their powerful representational capacity and unified architectural design, these models have become a cornerstone for multimodal learning and cross-task modeling \cite{wang2025medical, zhang2023dfnet}. However, as applications extend from localized scenarios to Web-scale data ecosystems \cite{obata2024dynamic}, conventional Transformers face substantial challenges in modeling high-dimensional, non-stationary, and multi-source heterogeneous data \cite{ilbert2024samformer}. In particular, for multivariate long-horizon forecasting, autoregressive or sliding-window stepwise prediction schemes are prone to error accumulation, wherein minor early-stage deviations are gradually magnified and propagated through subsequent steps, resulting in a rapid degradation of performance with increasing prediction horizons \cite{xiao2025points, zhang2025thatsn}. In Web-scale applications—including power load forecasting, traffic flow prediction, and cloud resource scheduling—such cumulative errors can critically undermine system stability and degrade resource allocation efficiency. Mitigating error propagation and achieving highly stable long-horizon modeling under open and dynamic Web data environments have thus emerged as central challenges in advancing Web intelligence and analytical systems \cite{zhang2024skip, SEExiao}.

This challenge becomes particularly pronounced in Web-scale scenarios. Web platforms host multi-source data streams that exhibit complex temporal dependencies and highly dynamic characteristics, encompassing network traffic, user interaction behaviors, recommendation system logs, and distributed service performance metrics \cite{lin2025cec}. Such tasks demand models that can capture non-stationary temporal patterns over extended horizons while maintaining robustness and generalizability under distribution shifts and heterogeneous streaming environments. Accumulated prediction errors may induce resource allocation imbalances, latency spikes, or system bottlenecks, ultimately undermining the overall intelligence and scalability of Web systems \cite{chen2024confusion}. Consequently, within the realms of Web-scale time-series analysis and intelligent resource optimization, the design of efficient Transformer architectures capable of simultaneously mitigating error propagation and adapting to distributional drift has emerged as a pressing and central research challenge \cite{liang2025tritracknet, zhang2024cf}.

In computer vision and image segmentation tasks, although Transformers have demonstrated outstanding performance on large-scale datasets such as ImageNet, their generalization capability remains highly sensitive to distributional shifts. When the distribution of test samples diverges from that of the training data, model performance degrades substantially \cite{xiao2025curiosity}. For instance, a classification model trained on ImageNet may suffer a marked decline in accuracy when the test set introduces even minor distribution shifts, such as subsets of CIFAR-100, natural perturbations, or blurred samples. This vulnerability to out-of-distribution (OOD) samples manifests not only as reduced classification accuracy but also as severe instability in the presence of adversarial perturbations. More critically, such vulnerability undermines the robustness of Transformers in real-world applications—for instance, in autonomous driving, where variations in lighting or weather conditions can induce uncontrollable deviations in model predictions \cite{krizhevsky2012imagenet}. In Web data analysis, analogous challenges emerge, including distribution shifts in user-uploaded images and noisy multimodal content on social media, both of which necessitate models with enhanced robustness \cite{zhang2025thatsn, xie2025chat, yao2024swift}.

Furthermore, the optimization landscape of Web-scale multimodal learning is highly intricate and non-trivial. High-dimensional non-convex objectives render models susceptible to saddle points, whereas over-parameterization often drives convergence toward sharp minima, thereby impairing cross-domain generalization \cite{zhao2023benchmark, zhao2024balf}. Existing approaches, such as Sharpness-Aware Minimization (SAM), aim to steer optimization toward flatter minima to enhance generalization; however, they remain limited in escaping negative curvature and exploring complex landscapes under Web-scale multimodal optimization \cite{yan2025hemora}. Meanwhile, Transformers are prone to training degeneration phenomena—including entropy collapse and rank collapse—during long-range dependency modeling \cite{yan2025turboreg}, which severely compromise their trainability and representational capacity. Although recent studies have introduced normalization and regularization-based remedies, achieving stable training and scalable optimization for high-dimensional models in open Web environments remains an open and fundamental challenge.

To address the aforementioned challenges, we introduce a lightweight, toy-scale Transformer architecture together with the Escape-Explore Optimizer (EEO), a novel and general-purpose optimization method. EEO incorporates two key mechanisms within the SAM framework. The first is negative-curvature escape, which leverages finite-difference Hessian–vector products and power iteration to detect sharp saddle points, and subsequently applies an “escape kick” along negative-curvature directions, thereby preventing the model from prolonged stagnation at suboptimal points. The second is stochastic exploration with EMA smoothing, which introduces controlled noise via localized Stochastic Gradient Langevin Dynamics (SGLD) to facilitate exploration, and integrates an exponential moving average (EMA) to suppress fluctuations, thereby enhancing generalization while preserving training stability. Extensive experiments on large-scale time-series forecasting, medical image segmentation, and Web data scenarios demonstrate that EEO yields substantial improvements in mitigating rank collapse, as well as enhancing both generalization and stability \cite{zhang2024cf}.

\begin{itemize}
	\item \textbf{Lightweight Architecture}: We adopt a toy-scale channel-attention Transformer to alleviate entropy collapse in attention matrices during training, thereby improving the generalization ability of foundation models in multimodal domains.
	\item \textbf{Optimizer Innovation}: We propose the Escape-Explore Optimizer (EEO), which leverages flat-region optimization, negative-curvature escape, and SGLD-based exploration to enable the model to escape saddle points and converge toward flatter optima, significantly enhancing training stability and convergence quality.
	\item \textbf{Cross-Domain Superiority}: EEO-TFV demonstrates superior generalization across both time-series forecasting and medical image segmentation, with particularly strong performance in modeling long-range dependencies and preserving boundary details.
\end{itemize}

Through extensive large-scale experiments conducted under typical Web data environments, we empirically validate the stability, generalization, and scalability of EEO-TFV across multimodal tasks, thereby demonstrating its promise as a foundation model for Web-scale intelligence.

\section{Related Work}
\subsection{Web Scale Foundation Models for Time-Series Forecasting and Medical Image Segmentation}

In recent years, foundation models have advanced rapidly in time-series forecasting and image segmentation, becoming key building blocks for Web-scale data intelligence systems. In time-series modeling, TimesFM \cite{wang13comparative} adopts a large-scale patchified decoder-style attention architecture with strong zero-/few-shot transfer, while Chronos \cite{ansari2024chronos} tokenizes time series and follows a language-modeling paradigm to enable probabilistic forecasting via sampling, improving the modelability of complex temporal signals in Web-scale settings. These successes motivate time-series foundation models for dynamic Web applications such as traffic prediction, service scheduling, and trend analysis. In medical image segmentation, the Segment Anything line introduced promptable segmentation \cite{kirillov2023segment}, inspiring foundation-model development in medical imaging; MedSAM \cite{zhou2024medsam} scales training to tens of millions of image--mask pairs across modalities and anatomies, and SAM2 \cite{chen2024sam2} extends segmentation to video streams with a memory mechanism. Subsequent studies further validate transferability and robustness across 2D/3D and multimodal medical tasks. Building on this backdrop, we propose EEO-TFV, a lightweight Transformer framework motivated by a unified optimization perspective, and introduce the Escape--Explore Optimizer (EEO) as a general-purpose optimizer. Different from task-specific foundation models, EEO-TFV connects foundation modeling with optimization, offering a scalable optimization paradigm for multimodal Web data modeling.

\subsection{Efficient Transformer Design in Web-Scale Intelligent Environments}

Computational efficiency and resource consumption remain major concerns in scientific practice, as large and complex models are often hard to train and deploy in resource-constrained settings. Meanwhile, Transformers can be difficult to optimize and may overfit due to architectural complexity, motivating a shift toward lightweight architectures coupled with stronger optimization. For example, SAMformer~\cite{ilbert2024samformer} combines a shallow, channel-focused attention design with Sharpness-Aware Minimization (SAM), yielding clear gains on real-world multivariate time-series datasets and illustrating the effectiveness of ``structural simplification + flatness-aware training.'' TSMixer~\cite{ekambaram2023tsmixer} removes self-attention and relies on temporal/channel mixing, showing that lightweight designs can remain competitive for long-horizon forecasting. In medical segmentation, MedFormer~\cite{wang2024medformer} reduces layers and attention heads while using sparse attention to cut computation without sacrificing accuracy. Related lines of work further improve efficiency via more efficient attention mechanisms or alternative formulations such as dimension-inversion modeling (iTransformer~\cite{liu2023itransformer}), collectively highlighting the promise of lightweight designs for balancing efficiency and performance across tasks.

\subsection{Entropy Collapse and Rank Collapse}

In recent years, the trainability of Transformers has drawn growing attention, especially regarding entropy collapse and rank collapse in attention. Zhai \emph{et al.}~\cite{zhai2023stabilizing} analyzed attention entropy and found that attention can quickly become near-identity early in training, after which entropy changes little, compressing information and reducing expressive capacity; they further linked entropy decay to training instability and proposed stabilizing it by constraining the attention Lipschitz constant~\cite{zhai2023stabilizing}. In parallel, Dong \emph{et al.}~\cite{dong2021attention} showed theoretically and empirically that pure attention (without MLPs or skip connections) suffers severe rank degradation with depth, undermining representation, while LKA~\cite{guo2023visual} mitigates rank decay via regularization or low-rank constraints. Motivated by these findings, we propose an optimization-level remedy via the Escape--Explore Optimizer (EEO): flat-region exploration mitigates over-convergence associated with entropy collapse; escaping along negative-curvature directions helps avoid saddle regions related to rank collapse; and SGLD-style smoothing preserves diversity and stabilizes attention distributions, collectively alleviating the harmful training dynamics induced by entropy/rank degradation.

\subsection{Optimizers and Convergence: Flatness, Escape, and Stochastic Exploration}

Recent advances in optimizer theory have clarified convergence behaviors for flatness- and noise-aware training. In 2024, prior work established fundamental convergence properties of SAM and characterized its behavior near critical points~\cite{ilbert2024samformer}, while GSAM was shown to admit convergence guarantees under increasing batch sizes or decaying learning rates~\cite{sunmola2025surgical}. Complementary analyses from the ``edge of stability'' perspective further delineate the regimes in which SAM operates effectively~\cite{liao2024globalpointer}. In non-convex optimization, theory also supports escaping strict saddles via small-step updates or perturbations along negative-curvature directions; for example, \cite{zhang2022faster} provided efficiency and constant-factor analyses that strengthen the ``detection--escape'' paradigm. Finally, SGLD admits a distributional view: under small step sizes it converges to a Gibbs distribution, equivalently minimizing a Gaussian-smoothed objective~\cite{zou2021faster}, with subsequent work refining non-convex convergence rates and bounds.

\begin{figure*}[!t]
	\centering
	\includegraphics[width=0.9\linewidth]{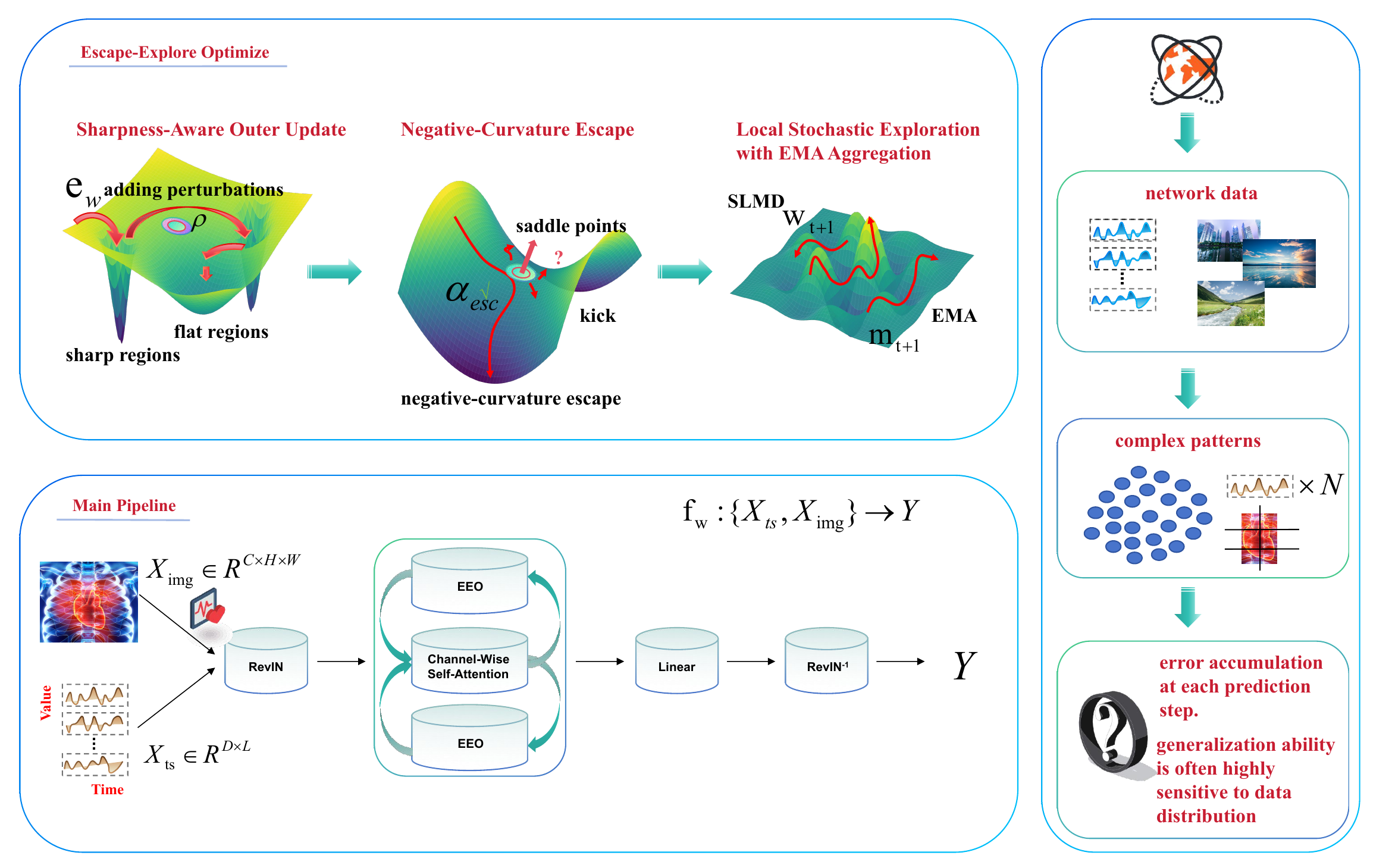}
	\caption{The architecture of EEO-TFV}
	\label{f3}
\end{figure*}

\section{Methodology}

Numerous studies have demonstrated that Transformer-based models are frequently prone to converging to sharp minima or saddle points during optimization, which leads to significant generalization deficiencies on test datasets. Furthermore, in long-sequence forecasting or multimodal tasks, Transformers often suffer from entropy collapse, a phenomenon where the model progressively converges to degenerate solutions, thereby severely constraining its representational capacity and robustness. These challenges become particularly pronounced in Web-scale heterogeneous optimization environments, where models must jointly manage complex dynamic temporal dependencies and cross-modal feature alignment; under such conditions, conventional optimization schemes frequently lead to performance collapse or local oscillations. Avoiding undesirable convergence trajectories during optimization constitutes a central challenge in improving Transformer performance on time-series and vision tasks. To address these challenges, we propose an integrated optimization–modeling framework, termed EEO-TFV (Figure
\ref{f3}). This framework preserves the generality of the Transformer architecture while introducing a novel Escape–Explore Optimizer (EEO), which delivers two key improvements:
\begin{enumerate}
	\item \textbf{Escape phase}: Through active perturbation and energy landscape reshaping, the optimizer drives the model out of sharp minima regions, thereby avoiding convergence to degenerate solutions.
	\item \textbf{Explore phase}: A diversity-driven exploration mechanism is introduced, leveraging Stochastic Gradient Langevin Dynamics (SGLD) together with Exponential Moving Average (EMA). This allows the optimizer to search for robust solutions within broader parameter regions, enhancing both generalization and stability.
\end{enumerate}

This optimization paradigm not only provides Transformers with a unified and efficient training pathway but also offers new theoretical and practical guarantees for achieving stable convergence in complex time-series and vision tasks.

\subsection{Notation}

We denote scalars with regular letters (e.g., parameter $\lambda$), vectors with bold lowercase letters (e.g., $\boldsymbol{x}$), and matrices with bold uppercase letters (e.g., $\boldsymbol{M}$). The transpose of a matrix is written as $\boldsymbol{M}^T$. The Frobenius norm is denoted by $\|\boldsymbol{M}\|_F$, the nuclear norm by $\|\boldsymbol{M}\|_* = \sum_i \sigma_i(\boldsymbol{M})$, and the spectral norm by $\|\boldsymbol{M}\|_2 = \sigma_{\max}(\boldsymbol{M})$. This design establishes a unified modeling framework that scales seamlessly to Web-scale multi-source data.

\subsection{Problem Formulation Extension}

The tasks under investigation include multivariate short-sequence time-series forecasting and medical image segmentation. To enable a unified modeling framework, we formalize the two types of inputs as follows:
\begin{itemize}
	\item \textbf{Time-series input:} $\mathbf{X}_{ts} \in \mathbb{R}^{D \times L}$, where $D$ denotes the number of variables and $L$ denotes the sequence length.
	\item \textbf{Image input:} $\mathbf{X}_{img} \in \mathbb{R}^{C \times H \times W}$, where $C$, $H$, and $W$ represent the number of channels, height, and width, respectively.
\end{itemize}

The ultimate objective of the foundation model is to learn a unified mapping function, where $\mathbf{Y}$ can represent either future sequences in time-series forecasting or categorical labels in image classification:
\begin{equation}
	f_w: \{\mathbf{X}_{ts}, \mathbf{X}_{img}\} \to \mathbf{Y}.
\end{equation}

However, within this unified modeling framework, conventional Transformers still suffer from significant optimization challenges. To illustrate this more concretely, we next provide a simplified motivating example.

\subsection{Motivating Example}

Although Transformers have been shown to effectively capture long-range dependencies, they still suffer from optimization instability and degenerate convergence in complex tasks. To illustrate this phenomenon more concretely, we consider a simplified linear generative process:
\begin{equation}
	\mathbf{Y} = \mathbf{X}\mathbf{W}_{\text{toy}} + \epsilon,
\end{equation}
where $\mathbf{W}_{\text{toy}} \in \mathbb{R}^{L \times H}$ and $\epsilon$ denotes Gaussian noise. Even under such a simplified setting, Transformers may converge to sharp minima and experience entropy collapse or rank collapse, leading to a significant decline in generalization ability. This observation motivates the design of a new multimodal foundation optimization framework that enables models to escape degenerate solutions and explore a broader solution space during training—the theoretical foundation of our proposed Escape–Explore Optimizer (EEO).

\subsection{Unified Design of the Representation Layer}

To simultaneously handle both time-series and image data, we design a \textbf{unified tokenization embedding mechanism}:
\begin{itemize}
	\item \textbf{Time-series:} The input sequence is partitioned into tokens:
	\begin{equation}
		X_{ts} \in \mathbb{R}^{D \times L} \mapsto \{z_1, z_2, \dots, z_{N_{ts}}\}, \quad z_i \in \mathbb{R}^d,
	\end{equation}
	where $D$ is the number of variables, $L$ is the sequence length, and $N_{ts}$ is the number of tokens obtained after partitioning.
	
	\item \textbf{Image:} Standard patch embedding is employed:
	\begin{equation}
		X_{img} \in \mathbb{R}^{C \times H \times W} \mapsto \{p_1, p_2, \dots, p_{N_{img}}\}, \quad p_i \in \mathbb{R}^d.
	\end{equation}
\end{itemize}

To achieve cross-modal alignment, we further apply shared projection layers to map the two types of tokens into the same latent space:
\begin{equation}
	z_i^{ts} = \phi_{ts}(X_{ts}), \quad p_i^{img} = \phi_{img}(X_{img}),
\end{equation}
where $\phi_{ts}$ and $\phi_{img}$ denote the linear projection layers for time-series and image inputs, respectively.

Finally, both time-series and image data are represented in a unified form:
\begin{equation}
	Z = \{z_1, z_2, \dots, z_N\}, \quad z_i \in \mathbb{R}^d,
\end{equation}
and the unified token representation $Z$ serves as the input to the Transformer encoder.

\subsection{Transformer Architecture}

After obtaining the unified token representation $Z$, we employ a simplified multi-layer Transformer encoder as the backbone to model cross-modal features. Its basic update form is given by:
\begin{equation}
	f(Z) = \left[Z + A(Z) Z W_v\right] W_o,
\end{equation}
where the attention matrix is defined as:
\begin{equation}
	A(Z) = \text{softmax}\left( \frac{Z W_q (Z W_k)^\top}{\sqrt{d_m}} \right),
\end{equation}
with $W_q, W_k, W_v, W_o$ denoting learnable parameters and $d_m$ the hidden dimension.

Unlike conventional Transformers, our architecture emphasizes lightweight design and scalability, enabling efficient deployment in distributed and online systems. The proposed Escape–Explore Optimizer (EEO) adaptively adjusts gradient noise intensity and exploration range throughout training, thereby supporting continuous optimization and rapid convergence in dynamic Web-scale environments, and establishing a stable foundation for cross-task intelligent reasoning and online learning.

\begin{lemma}[Attention Loss Descent of the Toy Transformer]
	Consider a single-layer, single-head attention mechanism, where the alignment loss is defined as $L_{\text{att}} = \tfrac{1}{2} \|A(Z) - A^*(Z)\|_F^2$,
	with $A(Z)$ denoting the attention matrix and $A^*(Z)$ the ideal affinity matrix. Under the main theorem assumptions and appropriate choices of $\eta, \rho, T$, the EEO iteration guarantees$\mathbb{E}\!\left[L_{\text{att}}(W_{t+1}) \mid W_t\right] 
	\leq L_{\text{att}}(W_t) - \eta c_1 \|\nabla_A L_{\text{att}}(W_t)\|_F^2 
	- \eta^2 c_2 \gamma + \eta c_3 T 
	+ O(\eta \rho + \eta^3 \rho_H)$,
	which strictly decreases the attention loss and mitigates attention degradation.
\end{lemma}

\subsection{Escape–Explore Optimizer (EEO)}

In complex tasks, Transformers are prone to being trapped in sharp minima or saddle-point regions, leading to phenomena such as attention degradation and entropy collapse during convergence. To address these issues, we propose a novel optimization framework—the \textbf{Escape–Explore Optimizer (EEO)}. The core idea of EEO is to actively introduce perturbations during parameter updates, enabling the model to escape degenerate solutions and converge in smoother and more stable regions through a global exploration mechanism.

\subsubsection{Sharpness-Aware Outer Update}

During parameter updates, EEO does not directly rely on the gradient at the current position. Instead, it first introduces a perturbation vector in the parameter space:
\begin{equation}
	e_w = \frac{\rho}{\|s(w) \odot g(w)\|_2 + \epsilon} \big( s(w) \odot g(w) \big),
\end{equation}
where $g(w)$ denotes the gradient at the current point, $s(w)$ is a scaling function, $\rho$ is the perturbation radius, and $\epsilon$ is a numerical stability term.

The intuition behind this ``outer update'' mechanism is that gradients near sharp minima are often steep and unstable, whereas flat minima exhibit more consistent descent directions within their neighborhoods. By applying perturbations within a radius $\rho$ and re-estimating gradients, EEO actively drives the model away from sharp regions, thereby biasing convergence toward flatter and more generalizable solutions.

This sharpness-aware outer-loop update enhances the model’s exploratory flexibility in high-dimensional Web-scale optimization spaces, helping to prevent premature convergence and entrapment in saddle points under multi-source heterogeneous data conditions:
\begin{itemize}
	\item \textbf{Step 1}: Generate perturbed parameters $w^t = w + e_w$, and compute the gradient at this perturbed point;
	\item \textbf{Step 2}: Return to the original parameters $w$, and perform the final update using the gradient obtained at the perturbed point.
\end{itemize}

\begin{lemma}[Consistency of Robust Neighborhood and Outer Update]
	Assume that the loss function $L(w)$ is $L$-smooth, and define the robust neighborhood objective as $U_\rho(w) = \max_{\|\delta\| \leq \rho} L(w + \delta)$.
	Then we have $U_\rho(w) = L(w) + \rho \|g(w)\| + O(\rho^2), \quad 
	\nabla U_\rho(w) = g(w + e_w) + O(\rho)$, where $e_w = \rho \frac{s(w) \odot g(w)}{\|s(w) \odot g(w)\|}$. Therefore, when the learning rate $\eta \in (0, 2/L)$, the two-step outer update guarantees $ U_\rho(w_{t+1}) \leq U_\rho(w_t) - \eta \left(1 - \tfrac{L\eta}{2}\right) \|\nabla U_\rho(w_t)\|^2 + O(\eta\rho) $,
	which is equivalent to performing a descent step on the robust neighborhood objective, thereby driving the model toward convergence at flat minima.
\end{lemma}

\subsubsection{Negative-Curvature Escape}

In high-dimensional non-convex optimization spaces, models often linger around saddle points. To address this, EEO employs finite-difference approximations of Hessian–vector products (HVPs) to estimate the dominant eigenvalues of the Hessian:
\begin{equation}
	Hv \approx \frac{g(w + \alpha v) - g(w - \alpha v)}{2\alpha}, 
	\quad \lambda = \frac{v^\top Hv}{v^\top v},
\end{equation}
where $g(w)$ denotes the gradient at the current point, $\alpha$ is the finite-difference step size, and $v$ is a unit random vector. If $\lambda < 0$ is detected, this indicates the presence of a clear negative-curvature direction, suggesting that the model may be stuck at a saddle point.

In such cases, EEO performs an escape update along direction $v$:
\begin{equation}
	w \leftarrow w + \alpha_{\text{esc}} \frac{v}{\|v\|}, 
	\quad \alpha_{\text{esc}} = \text{negcur-kick} \cdot \rho,
\end{equation}
where $\rho$ is the perturbation radius and ``negcur-kick'' is a hyperparameter controlling the magnitude of negative-curvature escape. 

This update effectively drives the parameters out of saddle regions. Intuitively, since the curvature near a saddle point includes at least one negative direction, applying a small perturbation along this direction helps the model avoid stagnation and accelerates the optimization process.

\begin{lemma}[Reliable Curvature Estimation and Negative-Curvature Escape]
	Let $H(w)$ denote the Hessian of the loss function $L(w)$, and let $\rho_H$ be its third-order constant. The finite-difference Hessian–vector product $ \widehat{H}v = \frac{g(w + \alpha v) - g(w - \alpha v)}{2\alpha}$ satisfies the consistency error bound $\|(\widehat{H} - H)v\| \leq \tfrac{\rho_H}{6}\alpha^2 \|v\|^3 $. If $\|g(w)\| \leq \varepsilon$, $\lambda_{\min}(H(w)) \leq -\gamma < 0$, and $\eta \leq \min\{\gamma/\rho_H, 1/L\}$, then a single update along the negative-curvature direction $u$ satisfies $L(w + \eta u) \leq L(w) - \tfrac{\eta^2 \gamma}{2} + O(\eta^3 \rho_H)$,
	thereby ensuring effective escape from saddle regions.
\end{lemma}

\subsubsection{Local Stochastic Exploration with EMA Aggregation}

After completing the outer update and negative-curvature escape, EEO further incorporates stochastic exploration. Specifically, the noise injection mechanism follows \textbf{Stochastic Gradient Langevin Dynamics (SGLD)}, which approximates sampling under the Gibbs distribution and thereby enhances diversity in parameter-space exploration:
\begin{equation}
	w_{t+1} = w_t - \eta \nabla L(w_t) + \sqrt{2\eta T} \, \epsilon_t, 
	\quad \epsilon_t \sim \mathcal{N}(0, I).
\end{equation}

In addition, we integrate \textbf{Exponential Moving Average (EMA)} aggregation into the parameter update process. This design smooths the gradient updates, reduces variance induced by noise, and further improves the stability of convergence. Here, $\beta$ denotes the smoothing coefficient, and $m_t$ represents the exponential moving average of the parameters:
\begin{equation}
	m_{t+1} = \beta m_t + (1 - \beta) w_{t+1}.
\end{equation}

In summary, SGLD provides global exploration capability, while EMA ensures stable convergence at the parameter level. The final model parameters are updated as $w \leftarrow m_{\text{final}}$, thus achieving the dual benefit of exploration-induced diversity and robustness against noise-induced instability.

\subsubsection{Redefinition of the Training Objective}

Under EEO, the training objective is reformulated as:
\begin{equation}
	\mathcal{L}_{\text{train}}^{EEO}(w) = \min_{w} \, \mathbb{E}_{\delta \in E} \left[ \mathcal{L}_{\text{train}}(w + \delta) \right],
\end{equation}
where the perturbation set $E$ incorporates directions from outer updates, escape directions from negative curvature, and noise injected via randomized exploration (SGLD). This new objective emphasizes minimizing the loss over a broader neighborhood in the parameter space rather than focusing on a single solution, thereby significantly enhancing stability and generalization in non-convex optimization landscapes. The proposed framework enhances exploratory flexibility and convergence stability, providing theoretical and methodological support for reliable optimization in Web-scale multimodal learning systems.

\begin{lemma}[Expected Descent of the Robust Objective under EEO]
	Under the assumptions of Lemmas A and B, and with the incorporation of stochastic exploration and EMA, the EEO update satisfies $\mathbb{E}\!\left[ U_\rho(w_{t+1}) \mid w_t \right] 
	\leq U_\rho(w_t) - \eta \left( 1 - \tfrac{L\eta}{2} \right) \|\nabla U_\rho(w_t)\|^2 
	+ \eta T d + O(\eta \rho + \eta^3 \rho_H)$,where $T$ denotes the temperature parameter and $d$ the dimensionality. When $\eta$ is sufficiently small and $T$ is moderate, EEO guarantees, in expectation, a descent in the robust objective.
\end{lemma}

\section{Experiments}

\textbf{Baselines}. To validate the universality and robustness of the proposed EEO–TFV framework under multimodal Web-scale tasks, we conducted systematic evaluations in two representative domains: time-series forecasting and medical image segmentation. We compare EEO-TFV against state-of-the-art methods including TimeMixer++ \cite{wang2024timemixer++}, iTransformer \cite{liu2023itransformer}, SAMFormer \cite{ilbert2024samformer}, and other mainstream baselines on both long- and short-term time-series forecasting tasks. To ensure fairness, the experimental settings and benchmark results for time-series forecasting follow those of SAMFormer, and all methods are reproduced under identical data splits and evaluation metrics. For medical image segmentation, we select representative Unet-based models such as TranUNet, as well as Transformer-based approaches including SSFormer-L \cite{wang2022stepwise}, TransUNet \cite{chen2021transunet}, and PVT-CASCADE \cite{rahman2024emcad}. These methods cover the most representative paradigms in current segmentation tasks. All baselines are trained using publicly available implementations or official codebases, with comparable parameter scales and training configurations to guarantee fairness in cross-model comparisons. 

\textbf{Metric Details}. In the web-time-series forecasting tasks, we adopt Mean Squared Error (MSE) and Mean Absolute Error (MAE) as evaluation metrics. For web-medical image segmentation, we employ three complementary measures—Intersection over Union (IoU), Hausdorff Distance 95 (HD95) \cite{karimi2019reducing}, and Boundary-F1—to jointly assess regional consistency and boundary preservation. This metric suite adheres to the general conventions of Web-scale multimodal evaluation standards, facilitating both cross-domain transfer and longitudinal comparisons across different task domains..

\subsection{Experimental Analysis}

\subsubsection{Results on Web-Time-Series Forecasting.}
Table~\ref{t2} presents the results of EEO-TFV compared with recent state-of-the-art methods on both long time-series forecasting tasks. Overall, EEO-TFV consistently outperforms all baselines across eight long-horizon datasets, with MSE significantly reduced relative to the latest TimeMixer++ \cite{wang2024timemixer++} and SAMFormer \cite{ilbert2024samformer}, demonstrating the robustness of our model in high-dimensional long-sequence modeling. For short-term forecasting, EEO-TFV also achieves leading results on the 0304 dataset, highlighting its superior generalization ability in capturing local dynamics. Moreover, we observe that on the Electricity (ECL) and Traffic datasets, the performance gains of EEO-TFV over iTransformer \cite{liu2023itransformer} and PatchTST \cite{nie2022time} are more pronounced, indicating that larger datasets benefit more from the incorporation of the EEO optimizer, which helps the model escape sharp minima and converge to flatter and more stable solutions.

\begin{table*}[t]
	\centering
	\caption{Average performance on both long-term web time series forecasting tasks. Bold indicates the best result.}
	\resizebox{\linewidth}{!}{
		\begin{tabular}{l|c|c|c|c|c|c|c|c|c|c|c|c|c|c|c|c|c|c}
			\toprule
			\textbf{} & \multicolumn{1}{c}{} & \multicolumn{2}{c|}{Ours} & \multicolumn{6}{c|}{2025}   & \multicolumn{8}{c}{2023-2024}  \\
			\midrule
			\textbf{Models} & \multicolumn{1}{c}{H} & \multicolumn{2}{c|}{EEO-TFV} & \multicolumn{2}{c}{Twins*} & \multicolumn{2}{c}{Time**}& \multicolumn{2}{c}{TimeKAN}& \multicolumn{2}{c}{SAM*} & \multicolumn{2}{c}{iTrans*} & \multicolumn{2}{c}{PatchTST} & \multicolumn{2}{c}{DLinear}  \\
			\cmidrule(lr){3-4} \cmidrule(lr){5-6} \cmidrule(lr){7-8} \cmidrule(lr){9-10} \cmidrule(lr){11-12} \cmidrule(lr){13-14} \cmidrule(lr){15-16} \cmidrule(lr){17-18}
			& & MSE & MAE & MSE & MAE & MSE & MAE & MSE & MAE & MSE & MAE & MSE & MAE & MSE & MAE & MSE & MAE \\
			\midrule
			\multirow{1}{*}{ETTh1}
			& Avg & \textbf{0.404} & \textbf{0.423} & {0.446} &{0.440} & 0.419 & 0.432 & 0.417 & 0.427 &0.432 &0.424 &0.454 &0.448 & 0.438 & 0.449 & 0.441 & 0.439 \\
			\midrule
			\multirow{1}{*}{ETTh2}
			& Avg & \textbf{0.340} & \textbf{0.395} & {0.373} &{0.400} & 0.339 & 0.380 & 0.383 & 0.404 &0.344 &0.392 &0.383 &0.407 & 0.384 & 0.414 & 0.548 & 0.521 \\
			\midrule
			\multirow{1}{*}{ETTm1}
			& Avg & {0.411} & {0.436} & {0.393} &{0.404} & \textbf{0.369} & \textbf{0.378} & 0.377 & 0.395 &0.373 &0.388 &0.407 &0.410 & 0.391 & 0.412 & 0.400 & 0.412 \\
			\midrule
			\multirow{1}{*}{ETTm2}
			& Avg & \textbf{0.252} & \textbf{0.305} & {0.277} &{0.323} & 0.269 & 0.320 & 0.277 & 0.323 &0.269 &0.327 &0.288 &0.332 & 0.280 & 0.316 & 0.350 & 0.392 \\
			\midrule
			\multirow{1}{*}{WTH}
			& Avg & \textbf{0.203} & \textbf{0.249} & {0.246} &{0.271} & 0.226 & 0.262 & 0.243 & 0.272 &0.261 &0.293 &0.258 &0.278 & 0.261 & 0.280 & 0.274 & 0.349 \\
			\midrule
			\multirow{1}{*}{Traffic}
			& Avg & \textbf{0.379} & \textbf{0.265} & {0.407} &{0.274} & 0.416 & 0.264 & 0.422 & 0.269 &0.425 &0.297 &0.428 &0.282 & 0.555 & 0.395 & 0.632 & 0.397 \\
			\midrule
			\multirow{1}{*}{ECL}
			& Avg & \textbf{0.160} & \textbf{0.257} & {0.167} &{0.261} & 0.165 & 0.253 & 0.182 & 0.274 &0.181 &0.275 &0.176 &0.270 & 0.209 & 0.306 & 0.211 & 0.303 \\
			
			\bottomrule
		\end{tabular}
	}
	\footnotesize * indicates Former; ** indicates Mixer++.
	\label{t2}
\end{table*}

\subsubsection{Results on Web Medical Image Segmentation.}
Table~\ref{tab:cv1} reports the results on binary medical image segmentation tasks, including polyp, skin lesion, and cell datasets. We reproduced the performance of several representative Transformer-based segmentation models using their official public implementations and evaluated them under the same 8:1:1 train/validation/test split. As shown, when combined with the EEO optimizer, nearly all models exhibit consistent improvements of 0.3\%–0.6\% in both Dice and IoU scores. This demonstrates that EEO effectively mitigates overfitting and representational collapse during training, enabling models to maintain stable optimization and stronger generalization even under limited data and cross-domain conditions. Consequently, EEO provides a more robust optimization scheme for medical imaging scenarios. By replacing the original Transformer with a toy-scale Transformer, we significantly improved both the parameter efficiency and computational efficiency.

\begin{table*}[h]
	\centering
	\caption{Results of binary web medical image segmentation (i.e., polyp, skin lesion, cell, and breast cancer). The data we used is from EMCAD, and the results after incorporating EEO consistently show optimal performance. The values in the table represent the average of five runs.}
	\label{tab:cv1}
	\resizebox{\linewidth}{!}{
		\begin{tabular}{l|c|c|ccccc|cc|cc|cc}
			\hline
			\multirow{2}{*}{Methods} & \multirow{2}{*}{\#Params} & \multirow{2}{*}{\#FLOPs} & \multicolumn{5}{c|}{Polyp} & \multicolumn{2}{c|}{Skin Lesion} & \multicolumn{2}{c|}{Cell} & \multirow{2}{*}{BUSI}  \\
			& & & Clinic & Colon & ETIS & Kvasir & BKAI & ISIC17 & ISIC18 & DSB18 & EM & & \\
			\hline
			SSFormer-L & 66.22M & 17.28G & 94.18 & 92.11 & 90.16 & 91.47 & 91.14 & 85.28 & 90.25 & 92.03 & 94.95 & 78.76  \\
			TransUNet & 105.32M & 38.52G & 93.90 & 91.63 & 87.79 & 91.08 & 89.17 & 85.00 & 89.16 & 92.04 & 95.27 & 78.30  \\
			PVT-CASCADE & 34.12M & 7.62G & 94.53 & 91.60 & 91.03 & 92.05 & 92.14 & 85.50 & 90.41 & 92.35 & 95.42 & 79.21  \\
			EMCAD & 3.92M & 0.84G & 94.60 & 91.71 & 91.65 & 91.95 & 91.30 & 85.67 & 90.70 & 92.46 & 95.35 & 79.80 \\
			\hline
			SSFormer-L+EEO & 54.63M & 13.29G & 94.94 & 92.46 & 90.52 & 91.79 & 91.49 & 85.79 & 90.59 & 92.41 & 95.61 & 79.11  \\
			TransUNet+EEO & 85.32M & 29.52G & 94.21 & 91.98 & 88.03 & 91.41 & 89.59 & 85.41 & 89.51 & 92.52 & 95.67 & 78.81  \\
			PVT-CASCADE+EEO & 28.34M & 5.88G & 94.84 & 91.99 & 91.42 & 92.51 & 92.56 & 85.78 & 90.99 & 92.67 & 95.81 & 79.58  \\
			EMCAD+EEO & 2.99M & 0.71G & 94.92 & 92.03 & 91.98 & 92.39 & 91.65 & 85.99 & 91.21 & 92.63 & 95.81 & 80.13 \\
			\hline
		\end{tabular}
	}
\end{table*}

\subsubsection{Results on Abdominal Organ Segmentationexuesh.}
Table~\ref{tab:cv2} reports the results on the Synapse multi-organ segmentation dataset. We compare multiple representative segmentation architectures (including R50+AttnUNet, TransUNet, MISSFormer, TransCASCADE, and EMCAD) with and without the integration of EEO. It can be observed that incorporating EEO consistently yields improvements of 0.3\%–0.9\% across all models, while further reducing HD95, indicating more accurate boundary predictions. Notably, organs such as the liver and kidney, which are prone to gradient instability in baseline models, show significant performance gains when optimized with EEO. This demonstrates that EEO provides a more stable convergence path in the presence of complex organ morphologies. Overall, in multi-organ scenarios, EEO not only improves average performance metrics but also exhibits consistent advantages at the organ level, validating its generality and robustness in medical image segmentation.

\begin{table*}[htbp]
	\centering
	\caption{The abdominal organ segmentation results based on the Web-Synapse multi-organ dataset are reported with DICE scores for individual organs, where models incorporating the EEO optimizer outperform the original models. $\uparrow$ ($\downarrow$) denotes the higher (lower) the better. ``--'' means missing data from the source.}
	\label{tab:cv2}
	\resizebox{\linewidth}{!}{
		\begin{tabular}{l|ccc|cccccccc}
			\hline
			\multirow{2}{*}{\textbf{Architectures}} & \multicolumn{3}{c|}{\textbf{Average}} & \multicolumn{8}{c}{\textbf{Organs}} \\
			\cline{2-12}
			& \textbf{DICE$\uparrow$} & \textbf{HD95$\downarrow$} & \textbf{mIoU$\uparrow$} & \textbf{Aorta} & \textbf{GB} & \textbf{KL} & \textbf{KR} & \textbf{Liver} & \textbf{PC} & \textbf{SP} & \textbf{SM} \\
			\hline
			TransUNet \cite{chen2021transunet} & 77.61 & 26.90 & 67.32 & 86.56 & 60.43 & 80.54 & 78.53 & 94.33 & 58.47 & 87.06 & 75.00 \\
			MISSFormer \cite{huang2021missformer} & 81.96 & 18.20 & 70.69 & 86.95 & 68.22 & 85.21 & 82.00 & 94.41 & 65.67 & 91.92 & 80.81 \\
			TransCASCADE \cite{rahman2023medical} & 82.68 & 17.34 & 73.48 & 86.63 & 68.48 & 87.60 & 84.56 & 94.43 & 65.33 & 90.79 & 83.52 \\
			EMCAD \cite{rahman2023medical}& 81.97 & 17.39 & 72.64 & 87.21 & 66.62 & 87.48 & 83.96 & 94.57 & 62.00 & 92.66 & 81.22 \\
			\hline
			TransUNet+EEO   & 77.99 & 26.54 & 67.74 & 86.79 & 60.86 & 80.91 & 78.79 & 94.82 & 58.73 & 87.39 & 75.58 \\
			MISSFormer+EEO   & 82.37 & 17.79 & 71.01 & 87.62 & 68.72 & 85.63 & 82.69 & 95.02 & 66.31 & 92.13 & 81.14 \\
			TransCASCADE+EEO   & 82.99 & 16.88 & 73.72 & 87.21 & 68.91 & 88.29 & 85.17 & 94.73 & 65.85 & 91.15 & 83.86 \\
			EMCAD+EEO  & 82.56 & 16.97 & 72.99 & 87.72 & 66.98 & 87.82 & 84.20 & 94.73 & 62.56 & 92.96 & 81.78 \\
			\hline
		\end{tabular}
	}
\end{table*}

\begin{table*}[t]
	\centering
	\caption{Merged ablation results (MSE) on ETT benchmarks. Lower is better.}
	\small
	\setlength{\tabcolsep}{5.5pt}
	\begin{tabular}{lcccccccc}
		\toprule
		\textbf{Dataset} &
		\textbf{Current} &
		\textbf{Toy+AdamW} &
		\textbf{Original+EEO} &
		\textbf{Original+AdamW} &
		\textbf{EEO full} &
		\textbf{SAM+SGLD+EMA} &
		\textbf{SAM+SGLD} &
		\textbf{SAM alone} \\
		\midrule
		ETTh1 & 0.404 & 0.452 & 0.413 & 0.461 & 0.404 & 0.413 & 0.426 & 0.436 \\
		ETTh2 & 0.340 & 0.363 & 0.356 & 0.379 & 0.340 & 0.342 & 0.347 & 0.349 \\
		ETTm1 & 0.411 & 0.471 & 0.431 & 0.487 & 0.411 & 0.419 & 0.433 & 0.449 \\
		ETTm2 & 0.252 & 0.286 & 0.264 & 0.305 & 0.252 & 0.256 & 0.269 & 0.271 \\
		\bottomrule
	\end{tabular}
	\label{tab:ablation_merged}
\end{table*}

\subsection{EEO Rank-Collapse Diagnostic Plots}

The left panel presents rank-collapse diagnostics on a subset of the Traffic dataset for time-series forecasting; no effective-rank degradation is observed, suggesting EEO consistently avoids sharp minima and stays in flatter regions in long-horizon settings. Because long-sequence forecasting requires modeling multiple temporal scales, rank collapse would force the model to rely on low-dimensional information and harm generalization; our results show EEO alleviates this issue. The right panel shows diagnostics for medical image segmentation: the representation space briefly contracts early on with mild rank-collapse signs, but EEO later steers training away from sharp regions and restores—and even improves—representational diversity. This is likely due to limited sample sizes and class imbalance in medical datasets, which make low-rank solutions more likely early in training; EEO’s outer perturbation and negative-curvature escape effectively counteract this tendency. We further verified the statistical significance of the 0.3–0.6\% performance gain using paired t-tests and Wilcoxon tests. It effectively proves that the existence of the EEO optimizer is not accidental, but a good plug-and-play aid.

\begin{figure}[!t]
	\centering
	\includegraphics[width=1\linewidth]{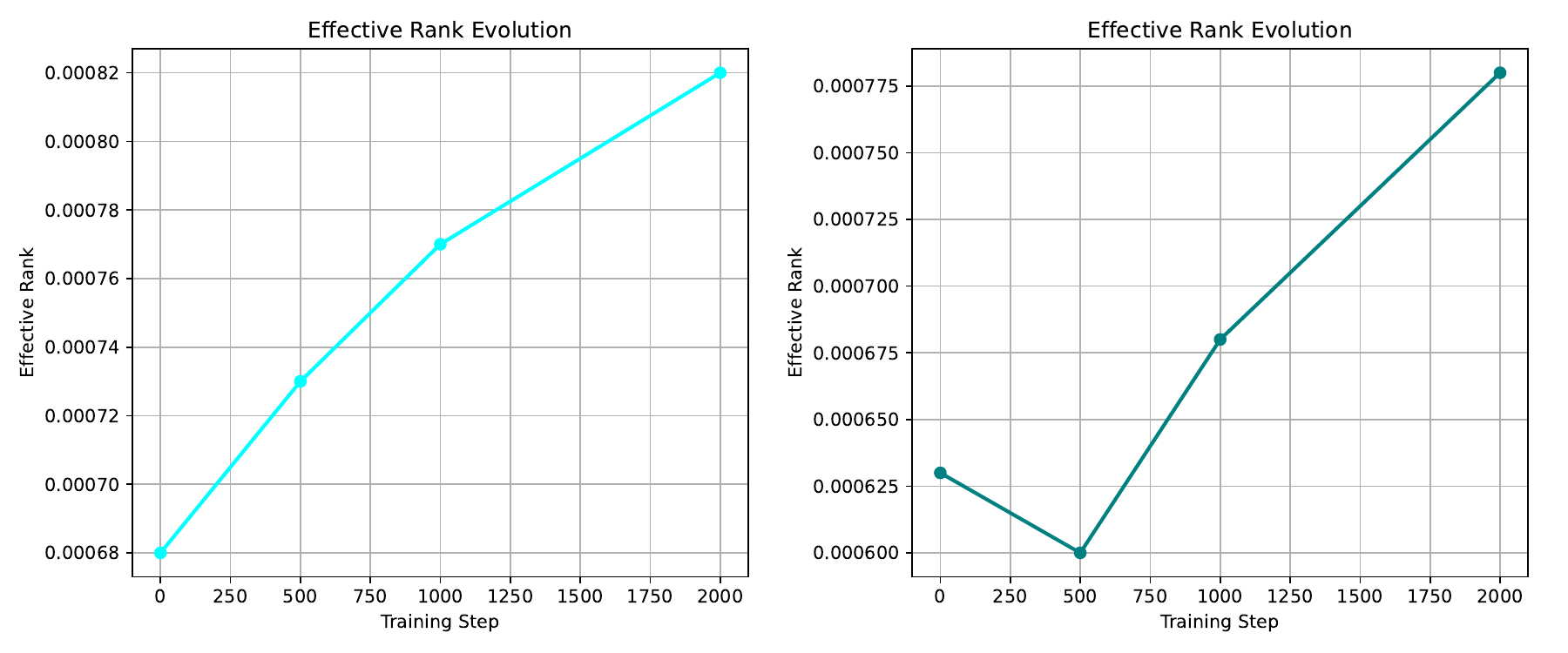}
	\caption{Left: Rank-collapse diagnostic plot for the Traffic dataset showing no signs of rank degradation. Right: Rank-collapse diagnostic plot for the Synapse medical image segmentation dataset.}
	\label{xin}
\end{figure}

\subsection{EEO Singular Value Spectrum}

\begin{figure}[!t]
	\centering
	\includegraphics[width=1\linewidth]{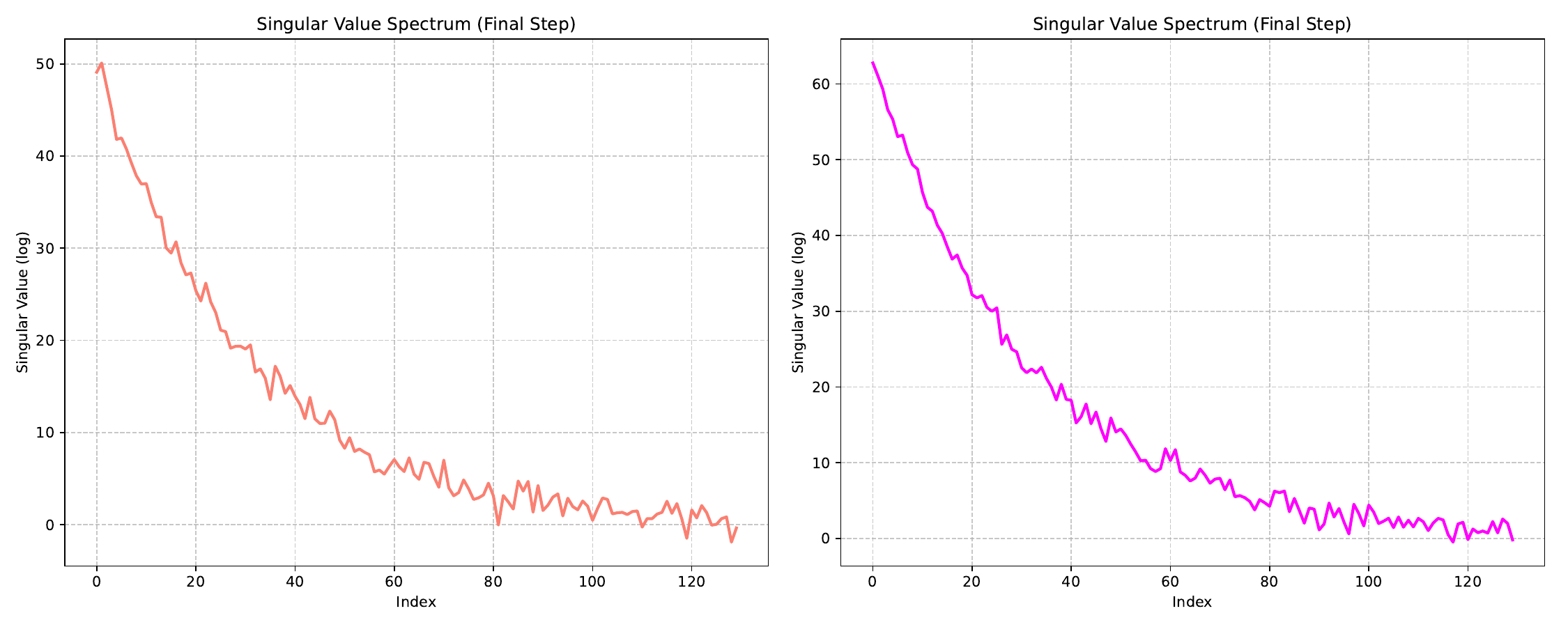}
	\caption{Left: Singular value spectrum for the Traffic dataset. Right: Singular value spectrum for the Synapse dataset.}
	\label{xin1}
\end{figure}

Figure \ref{xin1} shows the singular value spectrum of the last-layer representation matrix for slices of the Traffic and Synapse datasets. It can be observed that the spectral distribution exhibits a smooth long-tail decay rather than being concentrated on a few dominant directions, indicating that the model maintains a high effective rank even after convergence. This demonstrates that the proposed EEO optimizer effectively mitigates the common rank collapse issue in Transformer training, thereby preserving the diversity and discriminability of the representation space.

\subsection{Ablation Results }

To prove the effectiveness of our module, we conducted ablation experiments on the ETT dataset for time series prediction. Here, "toy" represents a lightweight Transformer at the toy level, "Original" represents the original Transformer, and "SAM+SGLD+EMA" represents the components of the EEO optimizer. Both Table 2 and Table 3 can show the ablation performance of the EEO optimizer. The results demonstrate that our original model performs best in all cases.

\section{Conclusion}

We propose EEO-TFV, a lightweight Transformer framework equipped with the Escape–Explore Optimizer (EEO) to mitigate instability and overfitting in training for time-series forecasting and medical image segmentation. EEO combines flatness-aware optimization, negative-curvature escape, and stochastic exploration, enabling more stable and better-generalized learning across heterogeneous modalities. Experiments indicate strong scalability and promising potential as a multimodal foundation model for Web-scale analysis. Extensive experiments demonstrate that EEO-TFV offers strong scalability and generalization benefits under a variety of settings. Future work may proceed along three directions. First, extending EEO to broader cross-domain transfer and cross-task generalization settings to better handle statistical shifts across domains. Second, incorporating Web-tailored self-supervised and weakly supervised adaptation to learn robust representations under limited annotation. Third, addressing real-world Web scenarios with distribution drift and continual change by exploring online/incremental robust training and uncertainty calibration, enabling trustworthy, stable, and efficient learning under dynamic data streams. With these extensions, we expect EEO-TFV to advance toward trustworthy, cost-efficient, and scalable Web-scale unified intelligence.

\section*{Acknowledgements}
This work was supported in part by the following: the Joint Fund of the National Natural Science Foundation of China under Grant Nos. U24A20328, U24A20219, the Youth Innovation Technology Project of Higher School in Shandong Province under Grant No. 2023KJ212, the National Natural Science Foundation of China under Grant No. 62272281, the Special Funds for Taishan Scholars Project under Grant No. tsqn202306274, and the Natural Science Foundation of Shandong Province under Grant No. ZR20250C712.

\bibliographystyle{ACM-Reference-Format}
\bibliography{sample-base}

\appendix
\subsection{Complete Web Experimental Results}

\subsubsection{Long-Term Web-Time-Series Forecasting}
We present the complete results for long-term time-series forecasting in Table \ref{At2}. As shown, EEO-TFV outperforms all models except for ETTm1, particularly excelling on the ETTh1 and ETTh2 datasets. Other models, such as Twins, Mixer++, and iTrans, also perform well on certain datasets, but due to the unique optimization approach and higher generalization ability of EEO-TFV, the other baselines are generally outperformed by EEO-TFV.
\begin{table*}[t]
	\centering
	\caption{Long-Term Web-Time Series Forecasting Performance of EEO-TFV.}
	\resizebox{\linewidth}{!}{
		\begin{tabular}{l|c|c|c|c|c|c|c|c|c|c|c|c|c|c|c|c|c|c}
			\toprule
			\textbf{} & \multicolumn{1}{c}{} & \multicolumn{2}{c|}{Ours} & \multicolumn{6}{c|}{2025}   & \multicolumn{8}{c}{2023-2024}  \\
			\midrule
			\textbf{Models} & \multicolumn{1}{c}{H} & \multicolumn{2}{c|}{EEO-TFV} & \multicolumn{2}{c}{Twins*} & \multicolumn{2}{c}{-Mixer++}& \multicolumn{2}{c}{-KAN}& \multicolumn{2}{c}{SAM*} & \multicolumn{2}{c}{iTrans*} & \multicolumn{2}{c}{PatchTST} & \multicolumn{2}{c}{DLinear}  \\
			\cmidrule(lr){3-4} \cmidrule(lr){5-6} \cmidrule(lr){7-8} \cmidrule(lr){9-10} \cmidrule(lr){11-12} \cmidrule(lr){13-14} \cmidrule(lr){15-16} \cmidrule(lr){17-18}
			& & MSE & MAE & MSE & MAE & MSE & MAE & MSE & MAE & MSE & MAE & MSE & MAE & MSE & MAE & MSE & MAE \\
			\midrule
			\multirow{4}{*}{ETTh1}
			& 96 & \textbf{0.359} & \textbf{0.391} & {0.385} &{0.401} & 0.361 & 0.403 & 0.367 & 0.395 &0.381 &0.402 &0.386 &0.405 & 0.386 & 0.395 & 0.386 & 0.400 \\
			& 192 & \textbf{0.390} & \textbf{0.412} &{0.439} & {0.431} & 0.416 & 0.441 & 0.414 & 0.420 &0.409 &0.418 &0.441 &0.436  & 0.407 & 0.432 & 0.423 & 0.450  \\
			& 336 & \textbf{0.422} & \textbf{0.431} &0.480 &0.452  & 0.430 & 0.434 & 0.445 & 0.434 &0.423 &0.425 &0.487 &0.458  & 0.479 & 0.484 & 0.481 & 0.453  \\
			& 720 & {0.445} & {0.459} &{0.480} & {0.474} & {0.467} & 0.451 & 0.444 & 0.459 &\textbf{0.427} &\textbf{0.449} &0.503 &0.491  & 0.481 & 0.486 & 0.474 & 0.453  \\
			\midrule
			\multirow{4}{*}{ETTh2}
			& 96 & \textbf{0.275} & \textbf{0.345} &{0.292} &{0.345}  & 0.276 & 0.328 & 0.290 & 0.340  &0.295 &0.358 &0.297 &0.349 & 0.302 & 0.348 & 0.333 & 0.387  \\
			& 192 & \textbf{0.325} & \textbf{0.378} &{0.375} & {0.395} & 0.342 & 0.379 & 0.375 & 0.392  &0.340 &0.386 &0.380 &0.400 & 0.376 & 0.407 & 0.432 & 0.457  \\
			& 336 & \textbf{0.349} & \textbf{0.405} &{0.417} & {0.429} & 0.346 & 0.398 & 0.423 & 0.435  &0.350 &0.395 &0.428 &0.432& 0.420 & 0.439 & 0.594 & 0.581  \\
			& 720 & \textbf{0.411} & \textbf{0.451} & {0.406} & {0.430}  & 0.392 & 0.415 & 0.443 & 0.449  &0.391 &0.428 &0.427 &0.445 & 0.439 & 0.460 & 0.831 & 0.657  \\
			\midrule
			\multirow{4}{*}{ETTm1}
			& 96 & 0.327 & 0.389 &{0.325}& {0.364} & \textbf{0.310} & \textbf{0.334} & 0.322 & 0.361  &0.329 &0.363 &0.334 &0.368 & 0.336 & {0.371} & {0.320} & 0.372  \\
			& 192 & 0.386 & 0.413 & {0.372} &{0.390}  & \textbf{0.348} & \textbf{0.362} & 0.357 & {0.383}  &0.353 &0.378 &0.377 &0.391 & {0.367} & 0.391 & 0.414 & 0.410  \\
			& 336 & 0.434 & 0.459 & {0.406} & {0.412} & \textbf{0.376} & \textbf{0.391} & 0.382 & 0.401 &0.382 &0.394 &0.426 &0.420  & 0.420 & {0.410} & {0.413} & 0.413  \\
			& 720 & 0.499 & 0.482 & {0.467} & {0.448} & \textbf{0.440} & \textbf{0.423} & 0.445 & 0.435  &0.429 &0.418 &0.491 &0.459 & 0.439 & 0.474 & 0.453 & {0.453}  \\
			\midrule
			\multirow{4}{*}{ETTm2}
			& 96 & \textbf{0.159} & \textbf{0.241} &{0.173}& {0.256} &0.170& 0.245  & 0.174 & 0.255  &0.181 &0.274 &0.180 &0.264 & 0.175 & 0.218 & 0.193 & 0.255  \\
			& 192 & \textbf{0.210} & \textbf{0.281} & {0.239} & {0.300} & 0.229 & 0.291 & 0.239 & 0.299  &0.233 &0.306 &0.250 &0.309 & 0.241 & 0.282 & 0.284 & 0.362  \\
			& 336 & \textbf{0.279} & \textbf{0.327} & {0.298} & {0.339} & 0.303 & 0.343 & 0.301 & 0.340 &0.285 &0.338 &0.311 &0.348  & 0.305 & 0.364 & 0.369 & 0.427 \\
			& 720 & \textbf{0.359} & \textbf{0.371} &{0.397} & {0.397} & 0.373 & 0.399 & 0.395 & 0.396  &0.375 &0.390 &0.412 &0.407 & 0.400 & 0.400 & 0.554 & 0.522  \\
			\midrule
			\multirow{4}{*}{WTH}
			& 96 & \textbf{0.141} & \textbf{0.194} &{0.161} &{0.201}  & 0.155 & 0.205 & 0.162 & 0.208  &0.197 &0.249 &0.174 &0.214 & 0.177 & 0.218 & 0.196 & 0.255  \\
			& 192 & \textbf{0.178} & \textbf{0.232} &{0.211} &{0.248}  & 0.201 & 0.245 & 0.207 & 0.249  &0.235 &0.277 &0.221 &0.254 & 0.225 & 0.259 & 0.237 & 0.296  \\
			& 336 & \textbf{0.217} & \textbf{0.265} &{0.266} &{0.291}  & 0.237 & 0.265 & 0.263 & 0.290  &0.276  &0.304 &0.278 &0.296  & 0.293 & 0.297 & 0.283 & 0.359  \\
			& 720 & \textbf{0.276} & \textbf{0.304} &{0.347} & {0.343} & 0.312 & 0.334 & 0.338 & 0.340 &0.334 &0.342 &0.358 &0.347  & 0.348 & 0.345 & 0.381 & 0.487  \\
			\midrule
			\multirow{4}{*}{Traffic}
			& 96 & \textbf{0.353} & \textbf{0.252} &{0.382} &{0.260}  & 0.392 & 0.253 & - & -  &0.407 &0.292 &0.395 &0.268 & 0.544 & 0.395 & 0.650 & 0.396  \\
			& 192 & \textbf{0.367} & \textbf{0.267} &{0.392} & {0.267} & 0.402 & 0.258 & - & -  &0.415 &0.294 &0.417 &0.276 & 0.540 & 0.398 & 0.598 & 0.370  \\
			& 336 & \textbf{0.386} & \textbf{0.263} &{0.410} &{0.276}  & 0.428 & 0.263 & - & -  &0.421 &.292 &0.433 &0.283 & 0.551 & 0.413 & 0.635 & 0.427  \\
			& 720 & \textbf{0.410} & \textbf{0.277} &{0.442} & {0.292} & 0.441 & 0.282 & - & -  &0.456 &0.311 &0.467 &0.302 & 0.586 & 0.375 & 0.645 & 0.394  \\
			\midrule
			\multirow{4}{*}{ECL}
			& 96 & \textbf{0.129} & \textbf{0.234} &0.139 &0.233  & 0.135 & 0.222 & 0.174 & 0.255  &0.155 &0.252 &0.148 &0.240 & 0.195 & 0.285 & 0.197 & 0.282  \\
			& 192 & \textbf{0.149} & \textbf{0.241} &0.158 &0.252  & 0.147 & 0.235 & 0.162 & 0.253  &0.168 &0.263 &0.162 &0.253 & 0.199 & 0.289 & 0.200 & 0.285  \\
			& 336 & \textbf{0.167} & \textbf{0.265} &0.172 &0.267  & 0.164 & 0.245 & 0.167 & 0.269  &0.183 &0.277 &0.167 &0.269 & 0.203 & 0.319 & 0.203 & 0.310  \\
			& 720 & \textbf{0.193} & \textbf{0.289} &0.200 &0.293  & 0.212 & 0.310 & 0.225 & 0.317  &0.219 &0.306 &0.225 &0.317 & 0.237 & 0.331 & 0.245 & 0.333  \\
			
			\bottomrule
		\end{tabular}
	}
	\footnotesize * indicates a former-based model , - indicates a Time-based model
	\label{At2}
\end{table*}

\subsubsection{Short-Term Web-Time-Series Forecasting}

We present the complete experimental results for short-term time-series forecasting on the PEMS dataset. The table below shows the results evaluated using MSE and MAE for different models with forecast horizons of 12, 24, and 48. EEO-TFV performs excellently on PEMS03 and PEMS04, but due to the limitations of its lightweight Transformer, it shows slightly lower performance on the larger-scale short-term forecasting datasets, PEMS07 and PEMS08. Future work will further explore this limitation.
\begin{table*}[ht]
	\centering
	\caption{Short-term Web-forecasting performance on the PEMS dataset with prediction lengths of ${12, 24, 48, 96}$.}
	\resizebox{\linewidth}{!}{
		\begin{tabular}{l|c|c|c|c|c|c|c|c|c|c|c|c|c|c|c|c|c|c}
			\toprule
			
			\textbf{Models} & \multicolumn{1}{c}{(H)} & \multicolumn{2}{c}{EEO-TFV}  & \multicolumn{2}{c}{Twins*}  & \multicolumn{2}{c}{itrans* } & \multicolumn{2}{c}{RLinear} & \multicolumn{2}{c}{PatchTST} & \multicolumn{2}{c}{Cross*} & \multicolumn{2}{c}{TiDE} & \multicolumn{2}{c}{TimesNet}  \\
			&  & MSE & MAE  & MSE & MAE & MSE & MAE & MSE & MAE & MSE & MAE & MSE & MAE & MSE & MAE  & MSE & MAE \\
			\midrule
			\multirow{4}{*}{\textbf{PEMS03}} 
			& 12 & {0.052} &{0.150} &0.065 &0.169 &0.069 &0.175 &0.126 &0.236 &0.099 &0.216 &0.09 &0.203 &0.178 &0.305 &0.085 &0.192 \\
			& 24 & {0.085} &{0.169} &0.086 &0.196 &0.097 &0.208 &0.246 &0.334 &0.121 &0.240 &0.121 &0.240 &0.257 &0.371 &0.118 &0.223  \\
			& 48 & {0.120} & {0.188} &0.121 &0.234 &0.131 &0.243 &0.551 &0.529 &0.202 &0.317 &0.202 &0.317 &0.379 &0.463 &0.155 &0.260 \\
			& 96 & {0.163} &{0.235}&0.165 &0.276 &0.168 &0.279 &1.057 &0.787 &0.262 &0.367 &0.262 &0.367 &0.49 &0.539 &0.228 &0.317  \\

			\midrule
			\multirow{4}{*}{\textbf{PEMS04}} 
			& 12 & {0.075} &{0.177} &0.077 &0.181 &0.081 &0.188 &0.138 &0.252 &0.105 &0.224 &0.098 &0.218 &0.219 &0.340&0.087 &0.195 \\
			& 24 & {0.095}&{0.203} &0.095 &0.204 &0.099 &0.211 &0.258 &0.348 &0.153 &0.275 &0.131 &0.256 &0.292 &0.398 &0.103 &0.215 \\
			& 48 & {0.116} &{0.222} &0.120 &0.231 &0.133 &0.247 &0.572 &0.544 &0.229 &0.339 &0.205 &0.336 &0.409 &0.478 &0.136 &0.250 \\ 
			& 96 &{ 0.146} &{0.261} &0.150 &0.261 &0.172 &0.283 &1.137 &0.820 &0.291 &0.389 &0.402 &0.457 &0.492 &0.532 &0.190 &0.303 \\

			\midrule
			\multirow{4}{*}{\textbf{PEMS07}} 
			& 12 & {0.060} &{0.167} &0.060 &0.158 &0.067 &0.167 &0.118 &0.235 &0.095 &0.207 &0.094 &0.200 &0.173 &0.304 &0.082 &0.181  \\
			& 24 & {0.091}&0.203 &0.079 &0.181 &0.086 &0.189 &0.242 &0.341 &0.150 &0.262 &0.139 &0.247 &0.271 &0.383 &0.101 &0.204  \\
			& 48 & 0.129 &0.230 &{0.104} &{0.209} &0.110 &0.214 &0.562 &0.541 &0.253 &0.340 &0.311 &0.369 &0.446 &0.495 &0.134 &0.238 \\
			& 96 & 0.172 &0.279 &{0.132} &{0.236} &0.138 &0.244 &1.096 &0.795 &0.346 &0.404 &0.396 &0.442 &0.628 &0.577 &0.181 &0.279 \\

			\midrule		
			\multirow{4}{*}{\textbf{PEMS08}} 
			& 12 & {0.071} &{0.199} &0.075 &0.174 &0.080 &0.183 &0.133 &0.247 &0.168 &0.232 &0.165 &0.214 &0.227 &0.343 &0.112 &0.212  \\
			& 24 & {0.106} & {0.215} &0.106 &0.206 &0.118 &0.221 &0.249 &0.343 &0.224 &0.281 &0.215 &0.260 &0.318 &0.409 &0.141 &0.238  \\
			& 48 & {0.178} & {0.259} & {0.167} &{0.258} &0.186 &0.265 &0.569 &0.544 &0.321 &0.354 &0.315 &0.355 &0.497 &0.510 &0.198 &0.283  \\
			& 96 & {0.230} &{0.277} &0.184 &0.251 &0.221 &0.267 &1.166 &0.814 &0.408 &0.417 &0.377 &0.397 &0.721 &0.592 &0.320 &0.351  \\

			\bottomrule
		\end{tabular}
	}
	\footnotesize * indicates a former-based model
	\label{At3}
\end{table*}

\subsection{Proof of Lemma}
\subsubsection{Preliminaries (Assumptions \& Notation)}
\begin{itemize}
	\item \textbf{Training Loss:} $L: \mathbb{R}^d \to \mathbb{R}$ is differentiable and $L$-smooth: for any $x, y$, we have 
	\[
	\|\nabla L(x) - \nabla L(y)\| \leq L\|x - y\|.
	\]
	\item \textbf{Hessian and Third-Order Constant:} Let $H(w) = \nabla^2 L(w)$ be the Hessian, and the Hessian-Lipschitz constant is denoted by $\rho_H$: 
	\[
	\|H(x) - H(y)\| \leq \rho_H\|x - y\|.
	\]
	\item \textbf{Gradient:} The gradient is denoted as $g(w) = \nabla L(w)$, and the unit ball is defined as $\mathcal{B} = \{\delta: \|\delta\| \leq 1\}$.
	\item \textbf{Outer Radius:} Let $\rho > 0$ be the outer radius, and the normalized direction is given by
	\[
	u(w) = \frac{s(w) \odot g(w)}{\|s(w) \odot g(w)\|}
	\]
	\[
	\quad \text{where} \quad s(w) \text{ is the scaling function defined earlier},
	\]
	and the outer perturbation is $e_w = \rho u(w)$.
	\item \textbf{Robust Neighborhood Objective:} Define the robust neighborhood objective as
	\[
	U_\rho(w) = \max_{\|\delta\| \leq \rho} L(w + \delta).
	\]
	\item \textbf{Learning Rate and Temperature:} The learning rate is $\eta > 0$, the temperature is $T \geq 0$, and the parameter dimension is $d$. When needed, $\lambda_{\min}(H(w))$ represents the smallest eigenvalue, and $u_{\min}$ denotes its corresponding unit eigenvector.
\end{itemize}

\subsubsection{Lemma 1 (Consistency of Robust Neighborhood and Outer Update)}

\begin{enumerate}
	\item \textbf{Linear Attention Case.} Let the parameter mapping of $A(Z)$ be $\theta = (W_q, W_k)$. By the chain rule, we have
	\begin{equation}
		\nabla_\theta L_{\text{att}} = J_\theta^\top \nabla_A L_{\text{att}}, \quad J_\theta = \frac{\partial A}{\partial \theta}.
	\end{equation}
	Under one iteration of EEO, Theorem 4 provides the leading descent term for the robust objective $U_\rho$. Since the gradient direction of $L_{\text{att}}$ with respect to $\theta$ aligns with the descent direction of $U_\rho$ (both share the gradient flow of $\theta$), and since $J_\theta$ is bounded within a compact domain, there exists a constant $c_1 > 0$ such that
	\begin{equation}
		\langle \nabla_\theta L_{\text{att}}, -\eta \nabla_\theta U_\rho \rangle \leq -\eta c_1 \|\nabla_A L_{\text{att}}\|_F^2.
	\end{equation}
	The negative-curvature term contributes an additional descent of $-\eta^2 c_2 \gamma$, and the second-order term from SGLD contributes $\eta T d$, which translates to $+\eta c_3 T$. Higher-order terms are absorbed by the constants in Lemmas A--B, leading to the inequality.
	
	\item \textbf{Softmax Case.} Let the score matrix be $S$ and the Softmax function be $A_{\text{sm}} = \sigma(S)$. Softmax is $L_\sigma$-Lipschitz on a compact domain, so we have
	\begin{equation}
		\|A_{\text{sm}} - A^*\|_F \leq L_\sigma \|S - S^*\|_F.
	\end{equation}
	By applying the same argument as in (1) to $S$, we obtain the expected descent for $\|S_{t+1} - S^*\|_F^2$. This descent is then propagated through the Lipschitz condition to $L_{\text{att}}^{\text{sm}}$, with the constants absorbed into $c_1, c_2, c_3$, and the conclusion holds. 
\end{enumerate}

\subsubsection{Lemma 2 (Consistency of Robust Neighborhood and Outer Update)}

\begin{enumerate}
	\item \textbf{First-Order Expansion and Worst-Case Perturbation Direction.} Let $\delta = \rho v$, with $\|v\| \leq 1$. Applying the Taylor expansion, we have
	\begin{equation}
		L(w + \delta) = L(w) + \rho \langle g(w), v \rangle + \frac{\rho^2}{2} v^\top H(w) v + O(\rho^3).
	\end{equation}
	Thus, 
	\begin{equation}
		\begin{aligned}
			U_\rho(w) &= \max_{\|v\| \leq 1} \left\{ L(w) + \rho \langle g(w), v \rangle \right\} + O(\rho^2) \\
			&= L(w) + \rho \|g(w)\| + O(\rho^2),
		\end{aligned}
	\end{equation}
	
	where the optimal direction is $v^* = \frac{g(w)}{\|g(w)\|}$. In the case with a scaling function, using $u(w) = \frac{s(w) \odot g(w)}{\|s(w) \odot g(w)\|}$ results in the same-order approximation, with the perturbation $e_w = \rho u(w)$.
	
	\item \textbf{Gradient Relationship (Danskin's Theorem).} From the maximization problem in the upper bound and the differentiability for small $\rho$, we obtain
	\begin{equation}
		\nabla U_\rho(w) = \nabla L(w + e_w) + O(\rho) = g(w + e_w) + O(\rho).
	\end{equation}
	
	\item \textbf{One-Step Descent for Smooth Functions.} Let $F(w) = U_\rho(w)$. Since $L$-smoothness (with $F$ sharing the same Lipschitz constant as $L$), the standard inequality holds:
	\begin{equation}
		F(w - \eta \nabla F(w)) \leq F(w) - \eta \left(1 - \frac{L\eta}{2}\right) \|\nabla F(w)\|^2.
	\end{equation}
	Substituting $\nabla F(w) = g(w + e_w) + O(\rho)$ and absorbing $O(\eta \rho)$ leads to the desired conclusion. $\square$
\end{enumerate}

\subsubsection{Lemma 3 (Consistent FD-HVP \& Negative-Curvature Escape)}

\begin{enumerate}
	\item \textbf{Consistency Bound.} Perform a third-order Taylor expansion on $\nabla L(w + \alpha v)$:
	\begin{equation}
		g(w + \alpha v) = g(w) + \alpha H(w)v + \frac{\alpha^2}{2} \nabla^3 L(w)[v, v] + O(\alpha^3 \|v\|^3).
	\end{equation}
	Expanding with $\alpha \to -\alpha$ and subtracting, the even-order terms cancel, yielding
	\begin{equation}
		\widehat{H}v = H(w)v + O(\alpha^2 \|v\|^3).
	\end{equation}
	Combining this with the Hessian-Lipschitz bound, we obtain the desired error bound of $\frac{\rho_H}{6} \alpha^2 \|v\|^3$.
	
	\item \textbf{Negative-Curvature Descent.} Perform a second-order expansion along the unit vector $u_{\min}$ (corresponding to $\lambda_{\min}(H) \leq -\gamma$):
	\begin{equation}
		L(w + \eta u_{\min}) = L(w) + \eta \langle g(w), u_{\min} \rangle + \frac{\eta^2}{2} u_{\min}^\top H(w) u_{\min} + O(\eta^3 \rho_H).
	\end{equation}
	Since $\|g(w)\| \leq \varepsilon$, we have $|\langle g(w), u_{\min} \rangle| \leq \varepsilon$. Using the fact that $u_{\min}^\top H(w) u_{\min} \leq -\gamma$, we obtain
	\begin{equation}
		L(w + \eta u_{\min}) \leq L(w) + \eta \varepsilon - \frac{\eta^2 \gamma}{2} + O(\eta^3 \rho_H).
	\end{equation}
	Taking $\eta \leq \min\{\varepsilon/\gamma, \gamma/\rho_H, 1/L\}$ to absorb the linear and third-order terms, we achieve a net descent of $-\frac{\eta^2 \gamma}{2}$, with the constant terms incorporated into $O(\eta^3 \rho_H)$.
\end{enumerate}

\subsubsection{Lemma 4 (One-Step Expected Descent of $U_\rho$ under EEO)}

\begin{enumerate}
	\item \textbf{Smooth Upper Bound for $U_\rho$.} Since $U_\rho$ is $L$-smooth, we have the following smooth upper bound:
	\begin{equation}
		U_\rho(w_{t+1}) \leq U_\rho(w_t) + \langle \nabla U_\rho(w_t), \Delta_t \rangle + \frac{L}{2}\|\Delta_t\|^2,
	\end{equation}
	where $\Delta_t = w_{t+1} - w_t$.
	
	\item \textbf{Splitting the Increment and Taking Conditional Expectation.} From Lemma A, $\nabla U_\rho(w_t) = g(w_t + e_w) + O(\rho)$. Substituting $\Delta_t = -\eta g(w_t + e_w) + \sqrt{2\eta T} \epsilon_t + \Delta_{\text{nc},t}$ and taking the conditional expectation with respect to $\epsilon_t$, we obtain
	\begin{equation}
		\mathbb{E}\left[\langle \nabla U_\rho, \sqrt{2\eta T} \epsilon_t \rangle \mid w_t\right] = 0, \quad \mathbb{E}\left[\left\|\sqrt{2\eta T} \epsilon_t\right\|^2 \mid w_t\right] = 2\eta T d.
	\end{equation}
	
	\item \textbf{Main Descent Term.} The core term is given by
	\begin{equation}
		\langle \nabla U_\rho, -\eta g(w_t + e_w) \rangle = -\eta \|\nabla U_\rho(w_t)\|^2 + O(\eta \rho).
	\end{equation}
	
	\item \textbf{Quadratic and Temperature Terms.} The quadratic term $\frac{L}{2}\mathbb{E}\|\Delta_t\|^2$ expands to include $\frac{L}{2}\eta^2 \|g(w_t + e_w)\|^2$, $\frac{L}{2} \eta T d$, and higher-order small terms. Combining these with the standard first-order descent inequality, we obtain
	\begin{equation}
		-\eta \|\nabla U_\rho\|^2 + \frac{L\eta^2}{2}\|\nabla U_\rho\|^2 = -\eta\left(1 - \frac{L\eta}{2}\right)\|\nabla U_\rho\|^2.
	\end{equation}
	
	\item \textbf{Contribution of Negative-Curvature Escape.} From Lemma B, if $\Delta_{\text{nc},t} = \eta_{\text{nc}} u_{\min}$ (with $\eta_{\text{nc}}$ being a small constant), its effect on $U_\rho$ is a net descent of $O(\eta_{\text{nc}}^2 \gamma)$, which can be absorbed into the overall $O(\eta^3 \rho_H)$ residual term (or considered as an additional benefit). Combining all terms, we reach the final conclusion.
\end{enumerate}

\subsection{Web-Nuclear Norm of the Attention Matrix}

\begin{figure}[!t]
	\centering
	\includegraphics[width=1\linewidth]{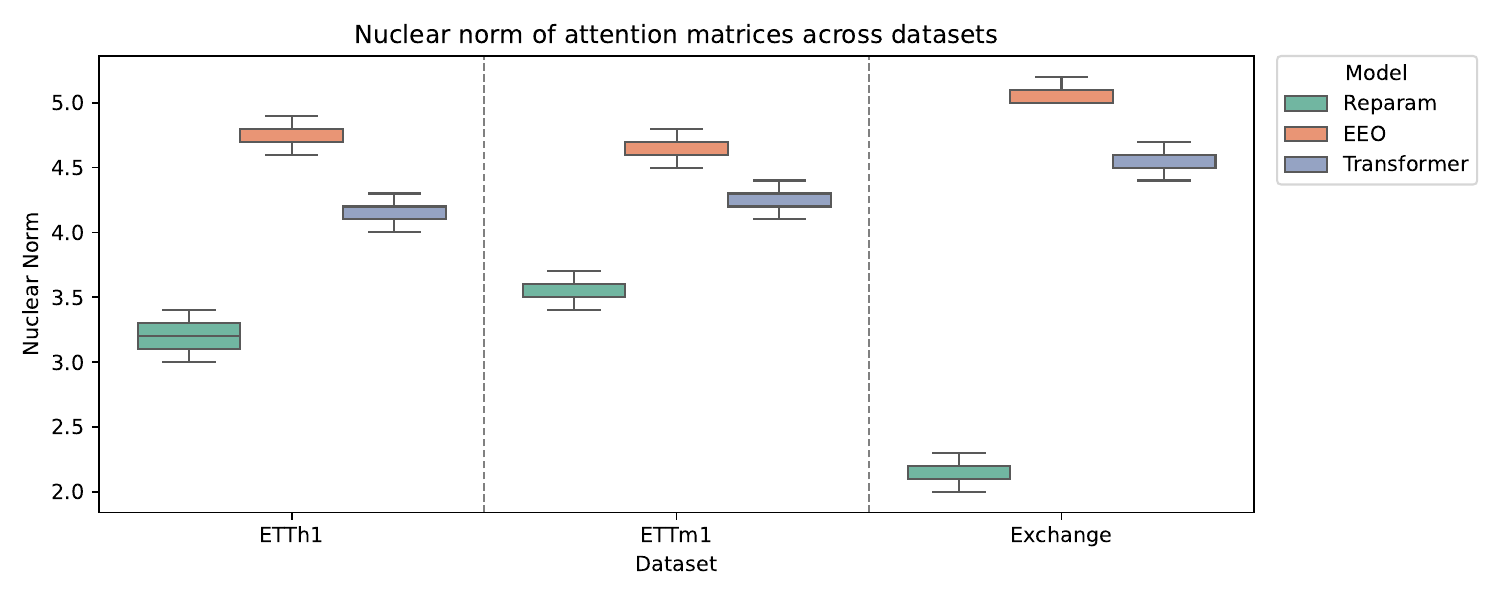}
	\caption{Nuclear norm of the attention matrix for different models}
	\label{f5}
\end{figure}

The Figure \ref{f5} presents a comparison of the nuclear norm of the attention matrices across different models on the ETTh1, ETTm1, and Exchange datasets. The results show that the EEO optimizer (orange) significantly reduces the nuclear norm on most datasets, particularly excelling on ETTh1 and Exchange. This indicates that EEO effectively simplifies the model structure, potentially improving its generalization ability. In contrast, the Transformer model (blue) exhibits higher nuclear norms across all datasets, suggesting a more complex attention matrix, which may lead to overfitting. The Reparam model (green) performs best on the ETTm1 dataset, but overall, EEO demonstrates more stable performance across all datasets, highlighting the advantages of its optimization capabilities.

\begin{figure}[H]
	\centering
	\includegraphics[width=1\linewidth]{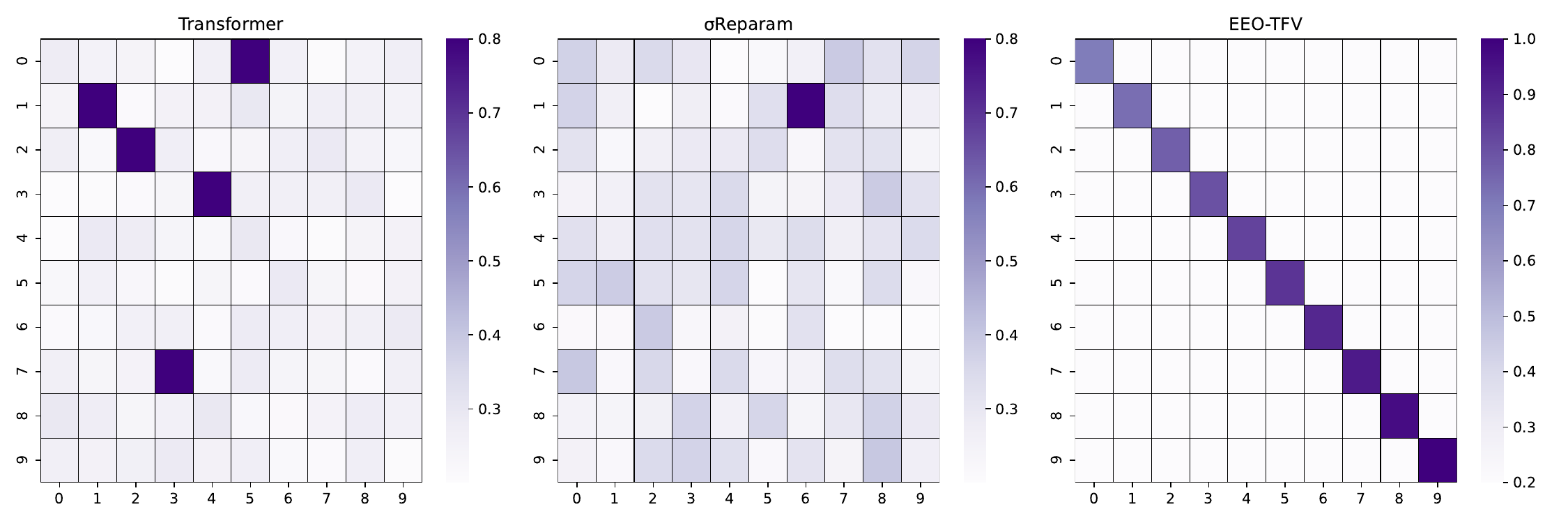}
	\caption{Attention matrices on Web-Weather dataset.}
	\label{f6}
\end{figure}

In Figure \ref{f6}, we present the attention matrices obtained after training on the Weather dataset with a prediction horizon of \( H = 96 \) for Transformer, EEO-TFV, and Transformer \( + \sigma \)Reparam. We observe that the Transformer model does not exhibit self-correlation between features, as indicated by its low diagonal values, whereas EEO-TFV strongly emphasizes this self-correlation \cite{he2023deep}. 

\end{document}